\documentclass{article}

\usepackage[preprint]{neurips_2025}

\usepackage[utf8]{inputenc} 
\usepackage[T1]{fontenc}    
\usepackage{hyperref}       
\usepackage{url}            
\usepackage{booktabs}       
\usepackage{amsfonts}       
\usepackage{nicefrac}       
\usepackage{microtype}      
\usepackage{xcolor}         
\usepackage{graphicx}       
\usepackage{amsmath}
\usepackage{enumitem}
\usepackage{bm}
\usepackage{pifont}

\newcommand{\methodname}{Budget Guidance}
\newcommand{\lowermethodname}{budget guidance}
\title{Steering LLM Thinking with Budget Guidance}

\author{%
  Junyan Li \\
  UMass Amherst\\
  \texttt{junyanli@umass.edu} \\
  \And
  Wenshuo Zhao \\
  Zhejiang University \\
  \texttt{zhao\_ws@zju.edu.cn} \\
  \And
  Yang Zhang \\
  MIT-IBM Watson AI Lab \\
  \texttt{yang.zhang2@ibm.com} \\
  \And
  Chuang Gan \\
  UMass Amherst \\
  \texttt{chuangg@cics.umass.edu}
}

\begin{document}

\maketitle

\begin{abstract}
    Recent deep-thinking large language models often reason extensively to improve performance, but such lengthy reasoning is not always desirable, as it incurs excessive inference costs with disproportionate performance gains. Controlling reasoning length without sacrificing performance is therefore important, but remains challenging, especially under tight thinking budgets. We propose \textit{budget guidance}, a simple yet effective method for steering the reasoning process of LLMs toward a target budget without requiring any LLM fine-tuning. Our approach introduces a lightweight predictor that models a Gamma distribution over the remaining thinking length during next-token generation. This signal is then used to guide generation in a soft, token-level manner, ensuring that the overall reasoning trace adheres to the specified thinking budget. \textit{Budget guidance} enables natural control of the thinking length, along with significant token efficiency improvements over baseline methods on challenging math benchmarks. For instance, it achieves up to a 26\% accuracy gain on the MATH-500 benchmark under tight budgets compared to baseline methods, while maintaining competitive accuracy with only 63\% of the thinking tokens used by the full-thinking model. \textit{Budget guidance} also generalizes to broader task domains and exhibits emergent capabilities, such as estimating question difficulty. The source code is available at: \url{https://github.com/UMass-Embodied-AGI/BudgetGuidance}.
\end{abstract}

\section{Introduction}
\label{sec:intro}

\begin{figure}[t]
    \centering
    \includegraphics[width=\textwidth]{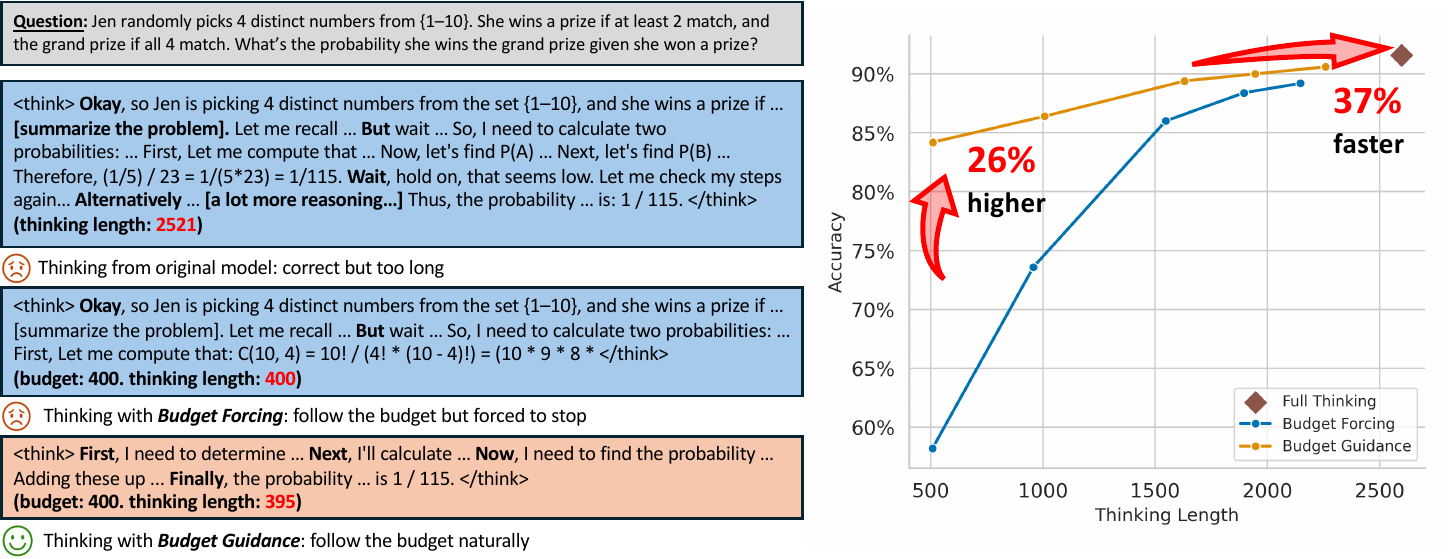}
    \caption{Deep-thinking models often produce excessively long reasoning traces, leading to high latency and unnecessary computation. Existing inference-time methods like \textit{budget forcing} rely on simplistic heuristics such as abruptly stopping, which can result in incomplete reasoning and degraded answer quality. In contrast, our method, \textit{{\lowermethodname}}, steers the reasoning process toward the target budget in a smoother and more natural way, without any LLM fine-tuning.}
    \label{fig:teaser}
\end{figure}

With the recent success of deep-thinking large language models (LLMs) -- such as OpenAI O1~\citep{jaech2024openai}, DeepSeek R1~\citep{guo2025deepseek}, and Qwen3~\citep{qwen2,qwen2.5}, which are capable of generating long sequences of thoughts to achieve better performance -- there has been a growing need to control the reasoning length of these models while maintaining the performance, because many deep-thinking LLMs often incur excessive inference costs with disproportionate performance gain. For example, in Figure~\ref{fig:teaser}, we show a response from a deep-thinking model that, while correct, is unnecessarily long. Such extensive reasoning is not always desirable, and there are cases where we need to impose a budget to limit the extent of reasoning, particularly in scenarios that demand real-time interaction, such as customer-facing chatbots, where excessive latency can degrade user experience and responsiveness.

Existing thinking budget control methods can be roughly divided into two categories with complementary strengths. The first category is fine-tuning methods, which fine-tune deep-thinking LLMs on specially curated dataset~\citep{han2024token} or with budget-aware reward to enable budget control capabilities~\citep{hou2025thinkprune}. Fine-tuning methods have been shown effective in changing the reasoning length while keeping competitive performance because they allow LLMs to fundamentally restructure and optimize their reasoning behavior according to the given budget. However, they come with two main drawbacks. First, fine-tuning an LLM is costly, requiring substantial computational resources and time. Second, directly fine-tuning the LLM may potentially alter its behavior in unexpected ways, such as compromising safety~\citep{qi2023fine}.

The second category of methods is the inference-time methods~\citep{ma2025reasoning,muennighoff2025s1}, which seek to alter the reasoning behavior at inference time. While these approaches do not involve fine-tuning, they often result in sub-optimal reasoning behaviors and significant performance degradation, because the intervention at inference time are often heuristic and overly simple, breaking the integrity of the original reasoning process. For example, one well-known inference-time method is \textit{budget forcing}~\citep{muennighoff2025s1} which terminates the model’s reasoning as soon as the thinking budget is reached, as described in Figure~\ref{fig:teaser}. While this method offers strict control over the number of generated tokens, abruptly interrupting the model may cut off unfinished thoughts and force premature answers, often leading to incorrect outputs.

In short, an important bottleneck in the task of thinking budget control lies in the tradeoff between \textit{non-intrusiveness} (in inference-time approaches) and \textit{optimality of the reasoning chain} (in fine-tuning approaches). This leads to our central research question: \textit{Can we design a flexible inference-time budget control approach (without fine-tuning) that still allows for wholistic, principled restructuring of the reasoning process to maintain its quality under budget?}

In this paper, we introduce \textit{{\lowermethodname}}, a novel approach that employs a lightweight auxiliary module to enable test-time control over the reasoning length of LLMs. Inspired by the principle of classifier guidance in diffusion models~\citep{dhariwal2021diffusion}, we train an auxiliary predictor that predicts the probability distribution of the remaining reasoning length at each reasoning step. The predicted length distribution is then used to modulate the LLM generation probability, effectively turning it into a budget-conditional generation probability. Our method avoids the direct fine-tuning of LLMs, while providing flexible and accurate control over the reasoning process. It can be seamlessly integrated into existing inference pipelines, and adapts to a wide range of models, thinking budgets, and tasks.

Our experiments have revealed several key highlights of our method. First, \textit{{\lowermethodname}} exhibits a remarkable trade-off between thinking length and performance. For example,  as shown in Figure~\ref{fig:teaser}, on MATH-500 benchmark~\citep{hendrycks2021measuring} \textit{{\lowermethodname}} can reduce the full thinking length by 37\% with minimal accuracy degradation, while been 26\% higher in accuracy than \textit{budget forcing} baseline under tight budget. Second, the auxiliary predictor is very successful in predicting the thinking length, effectively considering task difficulty and instruction type. Thus, it can accurately guide the thinking process under various budgets. Finally, our method demonstrates surprising generalizability across domains -- an auxiliary predictor trained on one dataset can also work well in other datasets and domains.

We summarize our contributions as follows:
\begin{itemize}[leftmargin=*, noitemsep, topsep=0pt]
  \item We propose \textit{\lowermethodname}, a novel test-time method for steering the reasoning process of LLMs toward a specified thinking budget, without requiring any fine-tuning of the LLM itself.
  \item We design a lightweight predictor that models a Gamma distribution over the remaining reasoning length based on the current generation context, and uses this signal to guide LLM generation toward a target thinking budget.
  \item \textit{Budget guidance} achieves strong trade-offs between thinking length and accuracy across multiple benchmarks, and demonstrates cross-domain generalization, enabling effective budget control and accurate thinking length prediction.
\end{itemize}

\section{Related Works}
\label{sec:related}

\subsection{Efficient LLM Reasoning}

Efficiency is a fundamental topic in machine learning, and recent work has focused on improving the token efficiency of LLMs in long chain-of-thought reasoning. For example, \textit{ThinkPrune}~\citep{hou2025thinkprune} employs reinforcement learning with an iterative pruning strategy to shorten reasoning traces; \textit{Z1}~\citep{yu2025z1} enables efficient test-time scaling by training on data with varying reasoning lengths and introducing a shifted thinking window for hybrid-mode inference; COCONUT~\citep{hao2024training} operates in continuous space to encourage reasoning with fewer tokens. While effective, these methods typically rely on expensive LLM fine-tuning and primarily aim to \textit{reduce} the length of reasoning, rather than to \textit{control} it.

More recent approaches~\citep{han2024token,muennighoff2025s1} have begun exploring methods to control the reasoning length, either through heuristic rules or model fine-tuning. In contrast, we propose a simple yet effective alternative: a fine-tuning-free approach that naturally steers the reasoning process to adhere to a specified thinking budget, enabling more efficient and flexible inference.

\subsection{Guidance and Guided Generation}

The term \textit{guidance} originates primarily from the diffusion model literature, where it denotes the ability to steer the generative process, often through truncated or low-temperature sampling, by reducing the variance or range of noise inputs to the generative model at sampling time~\citep{ho2022classifier}. This effectively transforms an unconditional diffusion model into a conditional one, enabling it to generate targeted outputs. One of the earliest examples is \textit{classifier guidance}~\citep{dhariwal2021diffusion}, which modifies the diffusion score by incorporating the gradient of the log-likelihood from an auxiliary classifier, thereby biasing the sampling process toward desired content. This can be viewed as a form of guided generation, where image generation is conditioned on the output of a classifier.

A similar notion of guided generation has emerged in the context of LLMs, where it typically refers to constraining the model's output to satisfy structural requirements, such as regular expressions or context-free grammars, to ensure syntactic correctness for downstream applications~\citep{willard2023efficient}.

To the best of our knowledge, our work is the first to extend the idea of guided generation to a new dimension: \textit{budget-conditioned generation}. Specifically, we introduce a novel form of guidance that softly steers the LLM’s generation to meet a specified thinking budget, enabling efficient and controlled reasoning without compromising output quality.

\section{{\methodname}}
\label{sec:method}

\begin{figure}[t]
    \centering
    \includegraphics[width=\textwidth]{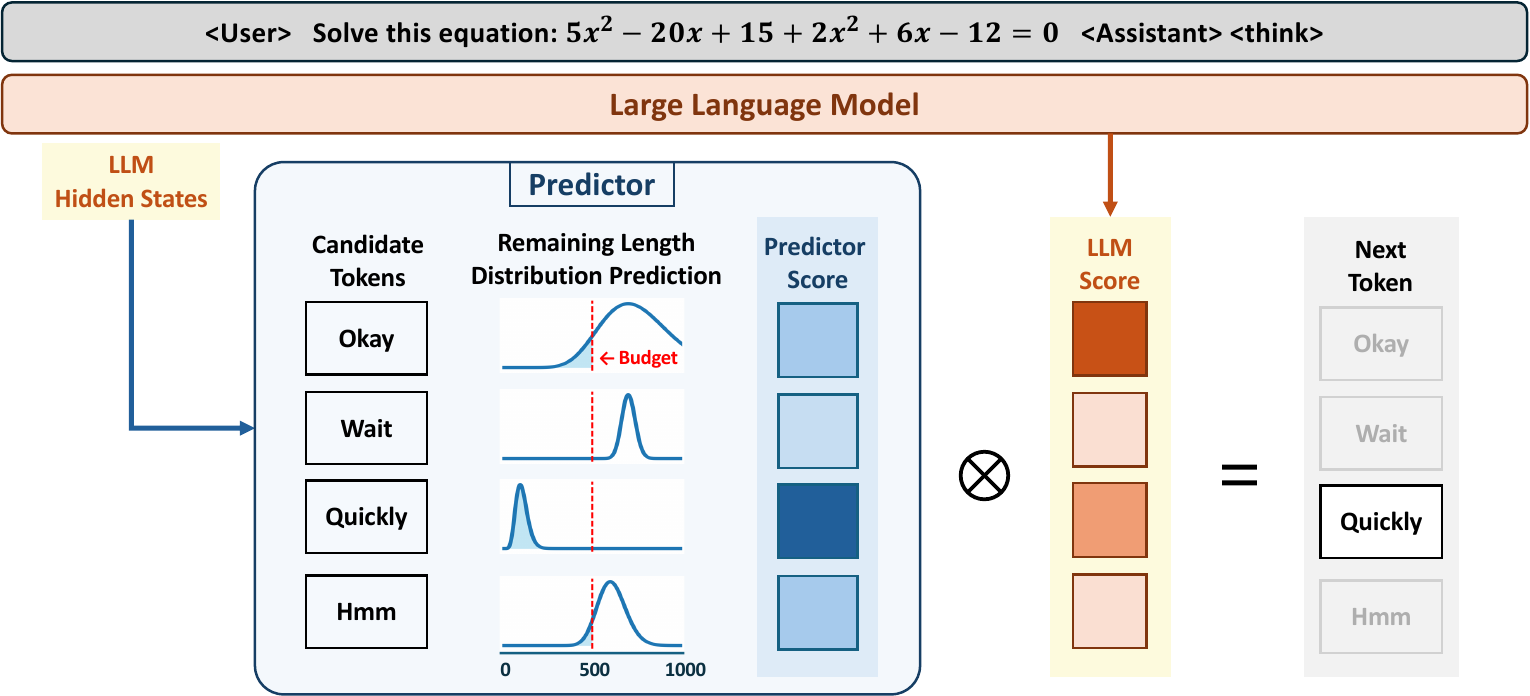}
    \caption{An overview of \textit{{\lowermethodname}}. A lightweight predictor uses the LLM’s hidden states to predict a Gamma distribution over the remaining reasoning length for each candidate token. We then use the CDF of Gamma distribution to compute a predictor score, which is combined with the LLM’s output score to guide generation. The result is soft, token-level steering toward budget-conditioned reasoning without any LLM fine-tuning.}
    \label{fig:method}
\end{figure}

We now introduce our method in detail. In Section~\ref{sec:budget-guidance}, we begin by formulating the budget-conditioned generation problem and present the overall \textit{budget guidance} framework, which draws inspiration from classifier guidance~\citep{dhariwal2021diffusion} in diffusion models. Section~\ref{sec:predictor} describes the design of our proposed auxiliary thinking length predictor, which estimates the distribution over remaining reasoning length at each decoding step. In Section~\ref{sec:training}, we outline the training procedure for the predictor using reasoning traces. Section~\ref{sec:architecture} introduces the model architecture of the predictor, which is designed to be lightweight and inference-efficient. Finally, Section~\ref{sec:skipping} presents a simple modulation-skipping strategy to further reduce computational overhead during decoding. An illustration of our method is provided in Figure~\ref{fig:method}.

\subsection{The Budget Guidance Framework}
\label{sec:budget-guidance}

The overall framework of our method follows the classifier guidance framework in diffusion generation~\citep{dhariwal2021diffusion}, thus we name our framework \textit{budget guidance}. Specifically, denote $X$ as the input question, $Y_{<t}$ as the LLM's output thinking process up to token $t$, and $Y_t$ as the LLM's output at token $t$. The LLM generation process essentially involves sampling from the following \textit{budget-unconditional distribution}, $p(Y_t | X, Y_{<t})$.

However, when there is a budget constraint, we would need to draw from a \textit{budget-conditional distribution}. Formally, denote $L_t$ as the random variable indicating the \textit{remaining length} of the thinking process from token $t$. For example, if the overall thinking length is $l$ (i.e., the \texttt{</think>} token occurs at token $l$), then $L_t = l-t$. Given the thinking budget limit $\bar{l}$, the budget-conditional distribution is defined as $p(Y_t | X, Y_{<t}, L_t \leq \bar{l}-t)$.

According to Bayes' rule, the budget-conditional distribution can be computed from the budget-unconditional distribution as follows
\begin{equation}
    \underbrace{p(Y_t | X, Y_{<t}, L_t \leq \bar{l}-t)}_{\text{budget-conditional}} \propto \underbrace{p(Y_t | X, Y_{<t})}_\text{budget-unconditional} \cdot Pr(L_t \leq \bar{l}-t | X, Y_{<t}, Y_t).
    \label{eq:bayes}
\end{equation}
Therefore, at each token $t$, generating from the budget-conditional distribution involves three steps. First, compute the unconditional distribution, which is simply performing a forward pass of the LLM. Second, predict the remaining length distribution, $Pr(L_t \leq \bar{l}-t | X, Y_{<t}, Y_t)$. Finally, use the remaining length distribution to modulate the unconditional distribution and then renormalize.

Therefore, within budget guidance framework, our task boils down to computing $Pr(L_t \leq \bar{l}-t | X, Y_{<t}, Y_t)$. To this end, we introduce a lightweight auxiliary thinking length predictor, which we describe in detail over the next three subsections.

\subsection{An Auxiliary Thinking Length Predictor}
\label{sec:predictor}

Denote the LLM vocabulary size as $n$, and denote the vocabulary as $\mathcal{V} = \{v_1, \ldots, v_n\}$. At each token $t$, the LLM outputs an $n$-dimensional unconditional probability vector (which we denote as $\bm u_t$):
\begin{equation}
    \bm u_t = [p(Y_t = v_1 | X, Y_{<t}), \ldots, p(Y_t = v_n | X, Y_{<t})].
\end{equation}
According to Equation~\eqref{eq:bayes}, the predictor needs to predict an $n$-dimensional vector (which we denote as $\bm a_t$):
\begin{equation}
    \bm a_t = [Pr(L_t \leq \bar{l}-t | X, Y_{<t}, Y_t = v_1), \cdots, Pr(L_t \leq \bar{l}-t | X, Y_{<t}, Y_t = v_n)],
    \label{eq:aux_vec}
\end{equation}
so that the budget-conditional probability vector, which we denote as $\bm c_t$, can be computed by element-wise multiplying the two vectors and renormalize: 
\begin{equation}
    \bm c_t = normalize( \bm u_t \circ \bm a_t).
    \label{eq:modulate}
\end{equation}

Equation~\eqref{eq:aux_vec} indicates that the predictor needs to accomplish a rather intensive task: At each token $t$, given the question $X$ and all the context generated so far $Y_{<t}$, the auxiliary predictor needs to \ding{182} traverse all possible values for $Y_t$ across the vocabulary, \ding{183} for each possible value, predict what would be the remaining length if $Y_t$ took on this value (that is $n$ probability distributions in total), and \ding{184} compute the cumulative probability up to $\bar{l}-t$ for each distribution.

To simplify the task, we parameterize each predicted distribution as a Gamma distribution for $\log(L_t)$:
\begin{equation}
    p(L_t | X, Y_{<t}, Y_t = v_i) = Gamma(\log(L_t); \lambda_{t}(v_i), \alpha_{t}(v_i)),
    \label{eq:gamma}
\end{equation}
where $Gamma(\cdot; \lambda, \alpha)$ represents the probability density function (PDF) of the Gamma distribution, with the shape parameter $\lambda$ and rate parameter $\alpha$. We model the distribution over $\log(L_t)$ instead of $L_t$ directly to better capture the dynamic range of thinking lengths.

With the Gamma distribution assumption, instead of predicting $n$ probability distributions, we only needs to predict two $n$-dimensional vectors: $\bm \lambda_t = [\lambda_{t}(v_1), \ldots, \lambda_{t}(v_n)]$ and $\bm \alpha_t = [\alpha_{t}(v_1), \ldots, \alpha_{t}(v_n)]$. The cumulative probability, $\bm a_t$, can be computed from the predicted $\bm \lambda_t$ and $\bm \alpha_t$ by the known closed-form cumulative distribution function (CDF) of the Gamma distribution.

\subsection{Training the Predictor}
\label{sec:training}

To train the predictor, we need to collect a dataset of reasoning chains produced by the target LLM. Formally, the data in the dataset takes the following form: $\mathcal{D} = \{(x,y_{1:l}, l)\}$, where $x$ is the input question, $y_{1:l}$ is the LLM-generated reasoning chain, and $l$ is the length of the reasoning chain. Note that the task dataset from which reasoning chain length training data are generated is not the same as the inference dataset (not even the same task), as we will show that the trained predictor has good dataset and task generalizability. For simplicity, in our training, we focus on math reasoning and use the OpenR1-Math-220k dataset~\citep{openr1}.

For each training datum, $(x,y_{1:l}, l)$, we feed the information of a partial reasoning chain to the predictor, truncated at different positions, and train the predictor to predict the remaining length. We adopt the maximum log likelihood objective for the gradient descent training. Formally, denote the parameters of the auxiliary predictor as $\bm \theta$. Then the training objective can be written as
\begin{equation}
    \max_{\bm \theta} \mathbb{E}_{(x, y_{1:l}, l)\sim \mathcal{D}} \bigg[\sum_{t=1}^{l-1} \log \big( p_{\bm \theta}(L_t=l-t | X=x, Y_{<t} = y_{<t}, Y_t = y_t) \big)\bigg],
\end{equation}
where $p_{\bm \theta}(\cdot)$ represents the predicted PDF by the auxiliary predictor, as shown in Equation~\eqref{eq:gamma}.

\subsection{Architecture of the Predictor}
\label{sec:architecture}

The predictor is designed to be lightweight enough to avoid significant computational overhead during decoding, yet expressive enough to capture both the input question and the ongoing reasoning context to produce a meaningful estimate of the remaining reasoning length. To this end, we adopt BERT-base~\citep{devlin2019bert} as the backbone of our predictor. Its input consists of the concatenated hidden states from all layers of the last generated token of the target LLM, which encode rich semantic information about both the input question and the reasoning history. A linear projection maps the LLM’s hidden dimensionality to the predictor’s input space, and a \texttt{[CLS]} token is used to summarize the hidden states. The final \texttt{[CLS]} representation is passed through another linear projection to produce an output matrix $M \in \mathbb{R}^{n \times 2}$, where each row corresponds to the parameters $\bm{\lambda}_t$ and $\bm{\alpha}_t$ of a Gamma distribution. A softplus activation~\citep{dugas2000incorporating} is applied to ensure both parameters are non-negative.

\subsection{Skipping Modulation}
\label{sec:skipping}

Ideally, probability modulation in Equation~\eqref{eq:modulate} would be applied at every decoding step $t$. To reduce computational overhead, however, we apply it only at the start of each reasoning paragraph, indicated by newline delimiters, where uncertainty is typically highest. The modulation is thus defined as:
\begin{equation}
    \bm c_t = 
    \begin{cases}
        \text{normalize}(\bm u_t \circ \bm a_t), & \text{if } t \text{ is the start of a reasoning paragraph} \\
        \bm u_t, & \text{otherwise}
    \end{cases}
\end{equation}

Empirically, we find that this modulation introduces only a 0.6\% increase in total latency for a 7B-parameter LLM, which is negligible in practice.



\section{Experiments}
\label{sec:exp}

\subsection{Settings}
\label{sec:settings}

\noindent\textbf{Training.} We apply our method to three deep-thinking models: \textit{DeepSeek-R1-Distill-Qwen-7B} (R1-7B)~\citep{guo2025deepseek}, \textit{DeepSeek-R1-Distill-Qwen-32B} (R1-32B)~\citep{guo2025deepseek}, and \textit{Qwen3-8B}~\citep{qwen2,qwen2.5}. Training is conducted on \textit{OpenR1-Math-220k}~\citep{openr1}, a dataset of 220k math problems from \textit{NuminaMath 1.5}~\citep{li2024numinamath} with reasoning traces generated by DeepSeek R1. We apply a simple data augmentation technique (detailed in the Appendix) to double the dataset size. During training, the LLMs are frozen and only the predictor is updated. We train for one epoch using a batch size of 8 and a constant learning rate of $1.0 \times 10^{-4}$ after warmup. Training takes 15 hours for R1-7B and Qwen3-8B, and 35 hours for R1-32B, using 8 NVIDIA H100 GPUs. All evaluations are conducted on the same hardware setup.

\noindent\textbf{Evaluation.} We evaluate our method on four representative math reasoning benchmarks: \textbf{MATH-500}~\citep{hendrycks2021measuring}, \textbf{AIME-2024}~\citep{aopsAIME}, \textbf{AMC}~\citep{aopsAMC12} (including both AMC12 2022 and AMC12 2023), and the math subset from \textbf{OlympiadBench}~\citep{he2024olympiadbench}. These benchmarks cover diverse mathematical topics, including arithmetic, algebra, combinatorics, \textit{etc.}, and span a broad range of difficulty levels.

Besides math benchmarks, we also extend our evaluation to broader domains to test the out-of-domain transferability of our math-data-trained predictor. Specifically, we further evaluate on \textbf{GPQA Diamond}~\citep{rein2024gpqa} for scientific reasoning, \textbf{FOLIO}~\citep{han2022folio} for logical reasoning, the numerical reasoning subset from \textbf{TableBench}~\citep{wu2025tablebench} for tabular numerical reasoning, and \textbf{LiveCodeBench}~\citep{jain2024livecodebench} (2024-08 - 2025-01 following~\citep{guo2025deepseek}) for code reasoning.

All experiments are conducted in a zero-shot manner, \textit{i.e.}, we do not perform further fine-tuning on the training sets of the evaluation benchmarks. We use greedy decoding for all evaluation.

\noindent\textbf{Baselines.} We compare our method with other methods that also do not finetune the LLM. Our main baseline is \textit{budget forcing}~\citep{muennighoff2025s1}, which enforces a hard token limit by appending an end-of-thinking delimiter 
(and optionally “Final Answer:”) 
to trigger early exit 
and force the model to produce its best guess. 
We use their open-sourced codebase for evaluation. We also include \textit{NoThinking}~\citep{ma2025reasoning} as a baseline, which bypasses the reasoning stage by inserting a fixed phrase as the thinking process: \texttt{Okay, I think I have finished thinking.} We also report results from the original model with full thinking as a reference.

\subsection{Main Results}

\subsubsection{Evaluation on Math Reasoning Benchmarks}

Since the predictor is trained on math data, we first evaluate its performance on math reasoning benchmarks to assess in-domain effectiveness. We set the thinking budget to approximately half the original model’s full thinking length and ensure the average thinking length (denoted as \#Tokens) comparable between our method and the baseline, and report the task accuracy.

Table~\ref{tab:math-results} summarizes the evaluation results on math reasoning benchmarks. Across all three models and four datasets, \textit{{\lowermethodname}} consistently outperforms \textit{budget forcing} under comparable average thinking lengths, effectively reducing the reasoning length without causing significant accuracy degradation. Compared to \textit{NoThinking}, \textit{{\lowermethodname}} achieves substantially higher performance, indicating that the reasoning traces are non-trivial and contribute meaningfully to task success.

These improvements are consistent across different model sizes (7B to 32B) and model families (DeepSeek vs. Qwen), highlighting the general applicability of our approach to diverse deep-thinking LLMs. Notably, even though the predictor for Qwen3-8B is trained on reasoning traces generated by DeepSeek-R1, it still performs well. This suggests that the training data can be \textit{model-agnostic}, provided the target LLM exhibits a similar reasoning style, for instance, using words like “wait” or “alternatively” to structure its reasoning process.

\begin{table}[t]
\small
\centering
\vspace{-3mm}
\caption{Evaluation results on math benchmarks.}
\begin{tabular}{lcccccccc}
\toprule
\textbf{} & \multicolumn{2}{c}{\textbf{MATH-500}} & \multicolumn{2}{c}{\textbf{AIME-2024}} & \multicolumn{2}{c}{\textbf{AMC}} & \multicolumn{2}{c}{\textbf{OlympiadBench}} \\
 & Acc. & \#Tokens & Acc. & \#Tokens & Acc. & \#Tokens & Acc. & \#Tokens \\ \midrule
 & \multicolumn{8}{c}{\textit{DeepSeek-R1-Distill-Qwen-7B}} \\
Thinking & 91.6 & 2598 & 36.7 & 4446 & 78.3 & 4338 & 56.9 & 3960 \\
NoThinking & 74.8 & - & 23.3 & - & 47.0 & - & 40.4 & - \\
Budget Forcing & 86.0 & 1547 & 16.7 & \textbf{2015} & 55.4 & 1872 & 47.1 & 1844 \\
\textbf{{\methodname}} & \textbf{88.2} & \textbf{1329} & \textbf{33.3} & 2046 & \textbf{60.2} & \textbf{1768} & \textbf{54.2} & \textbf{1755} \\ \midrule
 & \multicolumn{8}{c}{\textit{DeepSeek-R1-Distill-Qwen-32B}} \\
Thinking & 93.2 & 2226 & 72.6 & 7694 & 77.1 & 4156 & 61.9 & 3435 \\
NoThinking & 68.2 & - & 20.0 & - & 47.0 & - & 41.9 & - \\
Budget Forcing & 86.4 & 1525 & 40.0 & 2936 & 50.6 & 1567 & 50.7 & \textbf{1797} \\
\textbf{{\methodname}} & \textbf{90.0} & \textbf{1288} & \textbf{56.7} & \textbf{2873} & \textbf{69.9} & \textbf{1528} & \textbf{57.8} & 1820 \\ \midrule
 & \multicolumn{8}{c}{\textit{Qwen3-8B}} \\
Thinking & 96.2 & 4613 & 73.3 & 13660 & 91.6 & 8740 & 71.7 & 9424 \\
NoThinking & 84.8 & - & 33.3 & - & 56.6 & - & 53.5 & - \\
Budget Forcing & 90.2 & 2545 & 43.3 & 4010 & 77.1 & \textbf{3807} & 61.6 & 3712 \\
\textbf{{\methodname}} & \textbf{93.0} & \textbf{2062} & \textbf{50.0} & \textbf{3981} & \textbf{80.7} & 3869 & \textbf{65.6} & \textbf{3639} \\ 
\bottomrule
\end{tabular}
\label{tab:math-results}
\end{table}

\subsubsection{Accuracy–Thinking Length Tradeoff Analysis}

\begin{figure}[t]
    \centering
    \includegraphics[width=\textwidth]{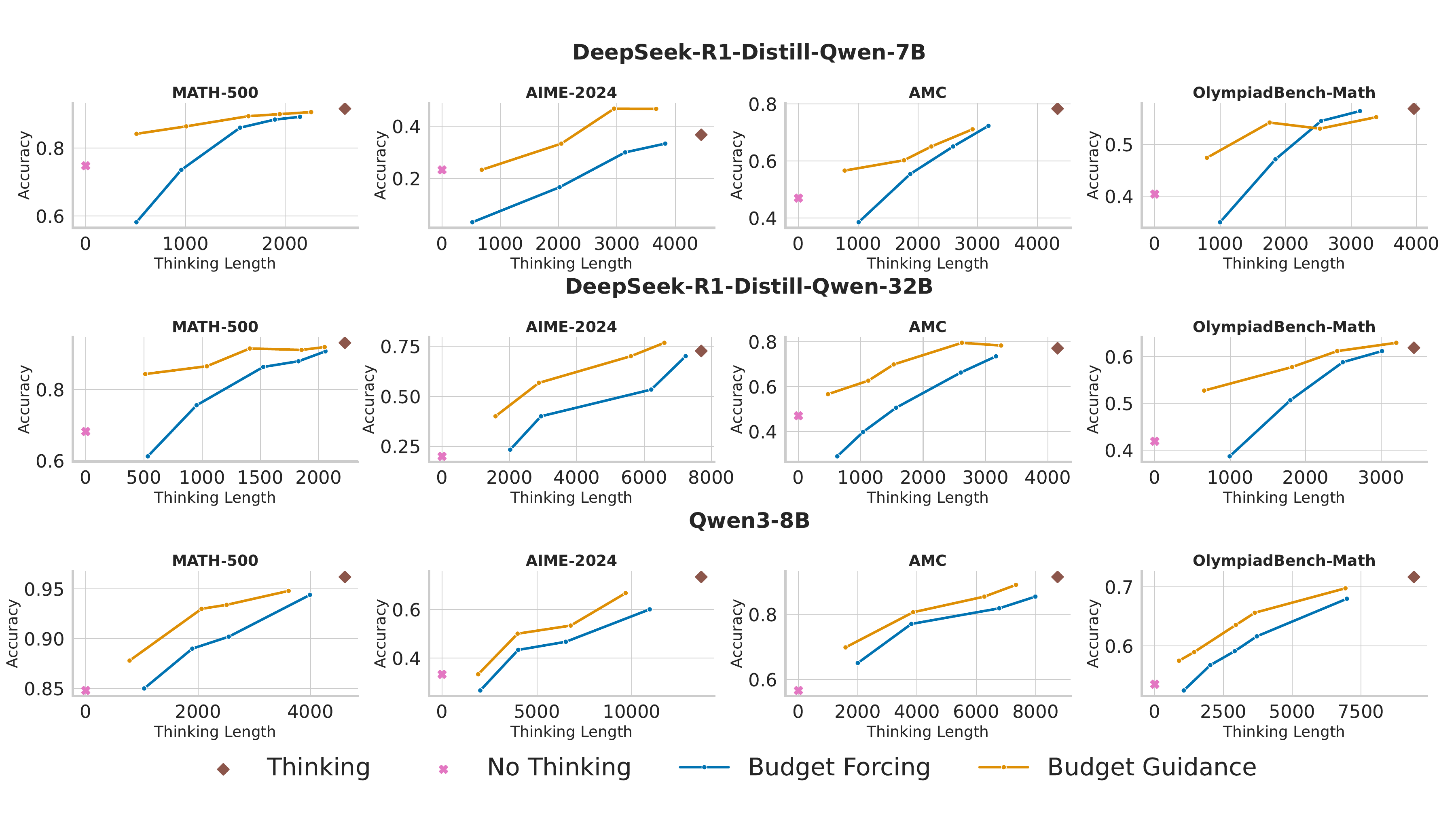}
    \caption{Accuracy vs. thinking length on math benchmarks.}
    \label{fig:all_acc}
\end{figure}

A key indicator of effective control is the ability to achieve higher accuracy under the same thinking length, which we call \textit{token efficiency}. To evaluate and compare the token efficiency of our method across different reasoning lengths, we vary the token budget to obtain different average thinking lengths and record the corresponding accuracy achieved by the model. We visualize this relationship through accuracy-thinking length trade-off curves. Experiments are conducted on all three models across the four math benchmarks, and the resulting plots are presented in Figure~\ref{fig:all_acc}.

From Figure~\ref{fig:all_acc}, we observe that our method consistently achieves better token efficiency across most benchmarks, achieving higher accuracy than \textit{budget forcing} under a range of thinking lengths. Notably, as the average thinking length decreases, corresponding to stricter budget constraints, our method yields significantly higher accuracy, particularly on benchmarks with diverse problem difficulty such as MATH-500. We attribute this to the ability of our method to adapt the reasoning pattern under strict budgets, producing concise yet complete reasoning traces. This enables the model to arrive at correct answers more efficiently, especially for questions that are relatively easy and do not require deep reasoning. This is also reflected in the occasional worse accuracy of \textit{budget forcing} compared to the \textit{NoThinking} baseline under strict budgets (\textit{e.g.}, MATH-500 on DS-7B/32B), where the reasoning trace is abruptly truncated and the model is forced to guess prematurely. In contrast, our method avoids such incomplete reasoning and consistently outperforms the \textit{NoThinking} baseline. An illustrative example of this guided reasoning behavior is provided in Section~\ref{sec:case_study}.

\subsubsection{Fine-Grained Control of Thinking Length}

\begin{figure}[t]
    \centering
    \includegraphics[width=\textwidth]{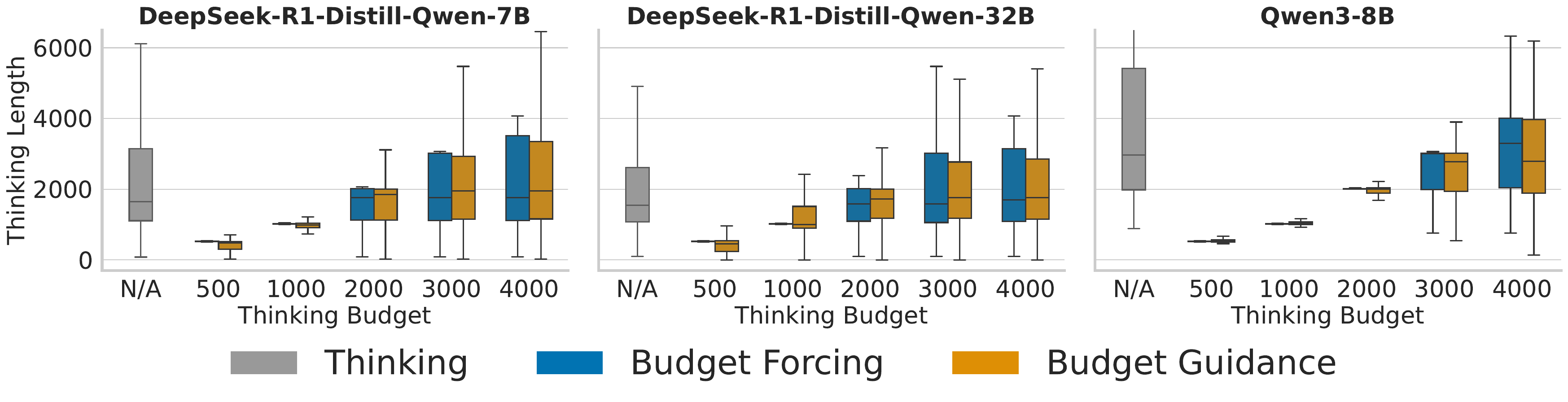}
    \caption{Thinking length controllability measured on MATH-500 benchmark.}
    \label{fig:control}
\end{figure}

Our goal is to steer LLM reasoning to adhere to a specified thinking budget. To evaluate controllability, we test on MATH-500 under varying thinking budgets, measuring the actual thinking length per sample and visualizing the distributions. We compare our method to \textit{budget forcing} and include the full-thinking baseline as a reference. Results across all three models are shown in Figure~\ref{fig:control}.

From Figure~\ref{fig:control}, we observe that our method behaves similar to \textit{budget forcing}, generally respects the specified thinking budget: for each setting, at least 75\% are within the budget, and the median thinking length closely aligns with the budget. Compared to the full-thinking baseline, our method guides the model to generate a budget-aligned reasoning trajectory. This behavior is notable because, unlike \textit{budget forcing}, our approach does not enforce a hard cutoff. Instead, it softly steers the generation process to match the desired level of detail, demonstrating flexible and controllable reasoning.

\subsubsection{Out-of-Domain Transferability}
\label{sec:ood}

\begin{table}[t]
\small
\centering
\caption{Evaluation on out-of-domain transferability.}
\begin{tabular}{@{}lcccccccc@{}}
\toprule
\textbf{} & \multicolumn{2}{c}{\textbf{GPQA Diamond}} & \multicolumn{2}{c}{\textbf{FOLIO}} & \multicolumn{2}{c}{\textbf{TableBench}} & \multicolumn{2}{c}{\textbf{LiveCodeBench}} \\
 & Acc. & \#Tokens  & Acc. & \#Tokens & Acc. & \#Tokens & Acc. & \#Tokens \\ \midrule
Thinking & 49.1 & 5838 &  63.5 & 849 & 37.0 & 906 & 26.9 & 3509 \\
NoThinking & 38.4 & - &  46.3  & - & 16.9 & - & 20.7 & - \\
Budget Forcing & 39.9 & 1895 &  60.1 & 372 & 22.4 & \textbf{379} & 28.8 & \textbf{1135} \\
\textbf{{\methodname}} & \textbf{49.0} & \textbf{1704} & \textbf{61.6} & \textbf{362} & \textbf{26.7} & 381 & \textbf{29.4} & 1138 \\
\bottomrule
\end{tabular}
\label{tab:out-of-domain-tasks}
\end{table}

While we train the predictor solely on math data for simplicity, we also explore its generalization to broader task domains. To this end, we conduct an out-of-domain transferability analysis using the DS-7B model. Specifically, we evaluate our method on four benchmarks: \textbf{GPQA Diamond} (scientific reasoning), \textbf{FOLIO} (logical reasoning), \textbf{TableBench} (tabular reasoning), and \textbf{LiveCodeBench} (code reasoning). We match the average reasoning length between our method and the baseline, and report the corresponding accuracies in Table~\ref{tab:out-of-domain-tasks}.

Despite being trained exclusively on math data, our predictor generalizes well to non-math reasoning tasks, consistently outperforming \textit{budget forcing} across all four benchmarks. These results highlight the cross-domain generalizability of our approach and its potential applicability to a wide range of reasoning scenarios. While the gains on out-of-domain tasks are less pronounced than those on in-domain benchmarks, we believe performance can be further improved by incorporating reasoning traces from a broader range of domains during training. We leave this direction for future work.

\subsection{Insights into What the Predictor Learns}
\label{sec:difficulty-proxy}

To probe what the predictor has learned, we analyze its estimated thinking length at the first thinking token, interpreted as the predicted number of thinking tokens needed, against \textbf{task difficulty} and \textbf{prompt type}, using the DS-7B model.

\begin{figure}[t]
  \centering
  \begin{minipage}[t]{0.55\textwidth}
    \centering
    \includegraphics[width=\linewidth]{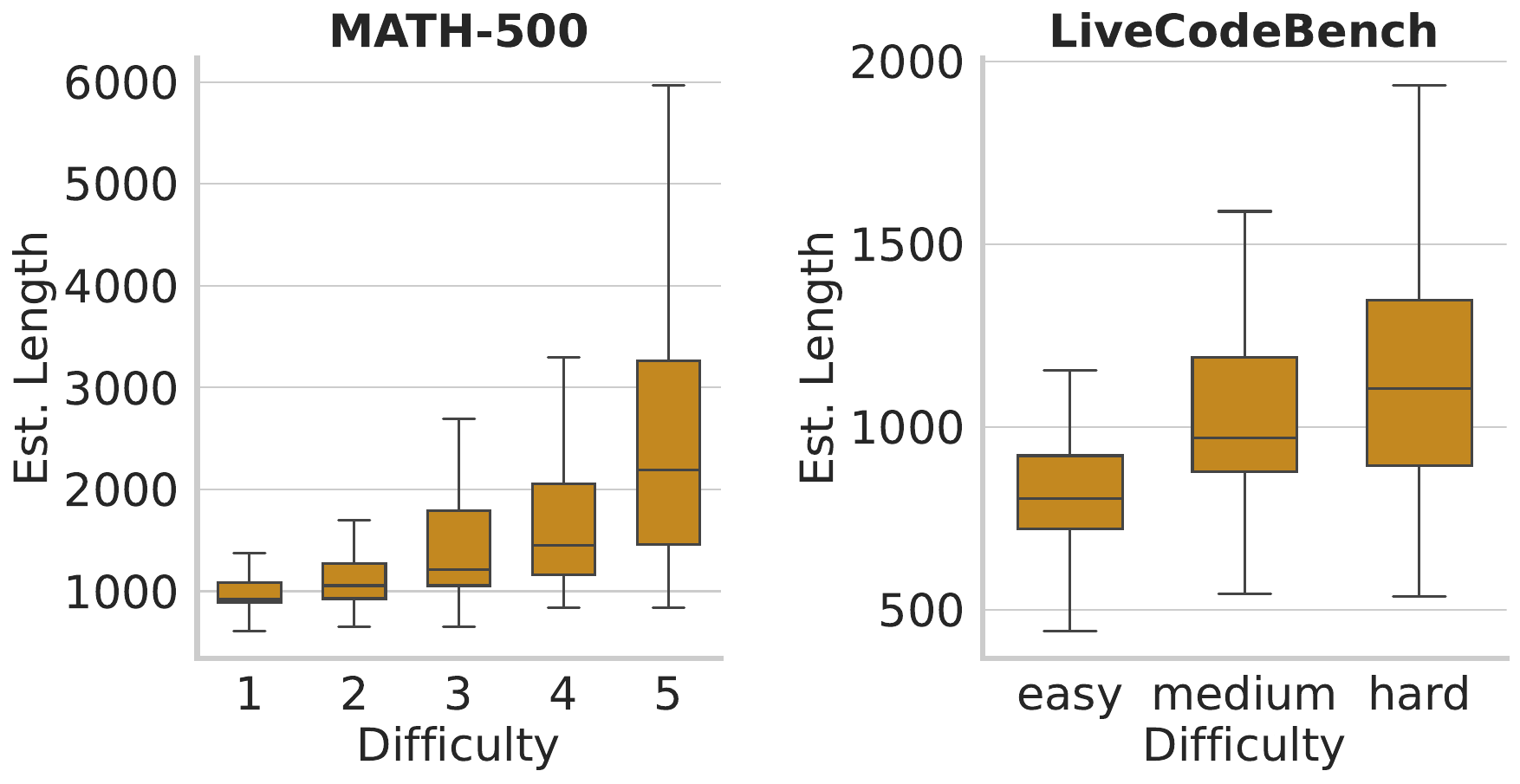}
    \caption{Correlation between question difficulties and estimated thinking lengths.}
    \label{fig:corr_diff}
  \end{minipage}
  \hfill
  \begin{minipage}[t]{0.4\textwidth}
    \centering
    \includegraphics[width=0.8\linewidth]{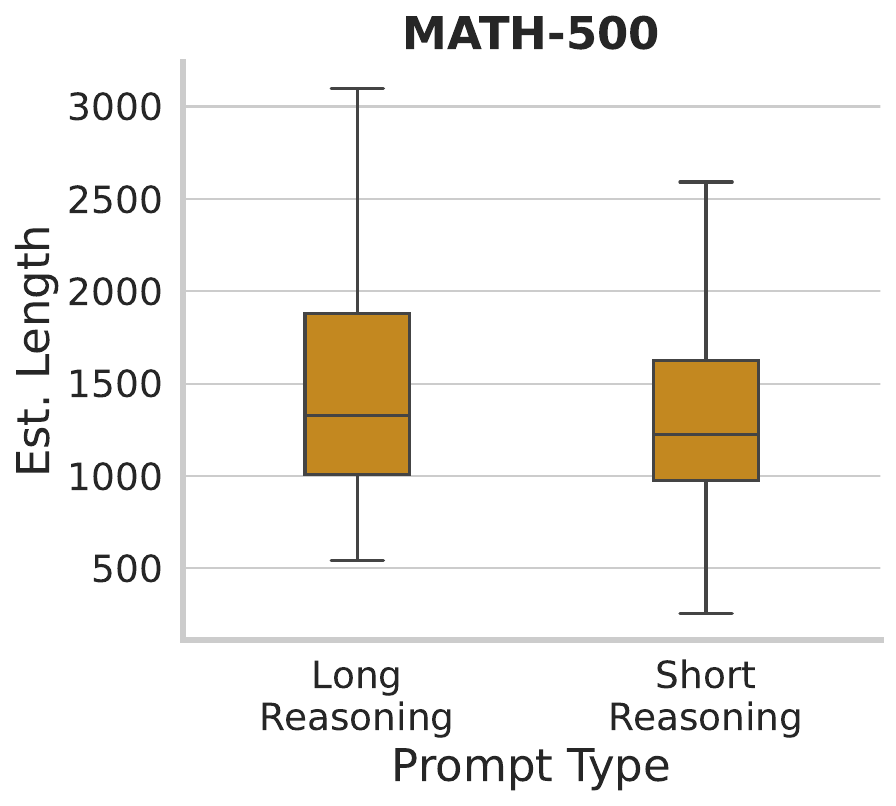}
    \caption{Correlation between prompt types and estimated thinking lengths.}
    \label{fig:corr_prompt}
  \end{minipage}
\end{figure}

\noindent\textbf{Task Difficulty.} We evaluate on MATH-500 (in-domain) and LiveCodeBench (out-of-domain). Figure~\ref{fig:corr_diff} shows that estimated thinking length increases with difficulty in both cases. This suggests that the predictor captures a general understanding of difficulty, enabling effective difficulty estimation.

\noindent\textbf{Prompt Type.} We evaluate on MATH-500 and compare two prompts: one encouraging long reasoning and one encouraging concise reasoning (listed in the Appendix). As shown in Figure~\ref{fig:corr_prompt}, the long reasoning prompt yields longer estimated thinking lengths. A t-test gives a $p$-value of 0.0028, confirming the difference is statistically significant and indicating that the predictor is prompt-aware.

\subsection{Case Study}
\label{sec:case_study}

\begin{figure}[t]
    \centering
    \includegraphics[width=\textwidth]{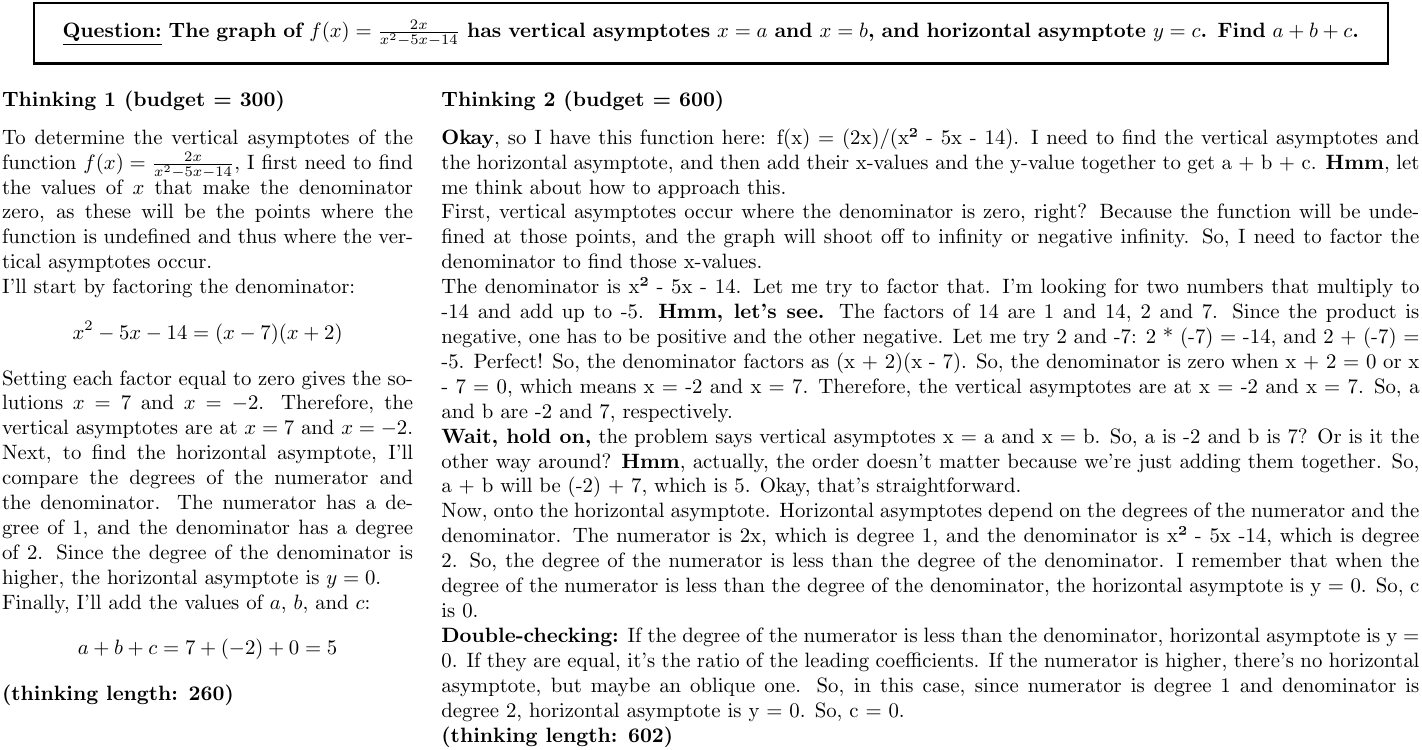}
    \caption{Sample reasoning traces generated with \textit{{\lowermethodname}} under different thinking budgets.}
    \label{fig:case_study}
\end{figure}


Figure~\ref{fig:case_study} shows a case study from MATH-500 illustrating reasoning traces under different thinking budgets. Rather than truncating output, our method adapts the reasoning style to the budget. With a stricter budget (left), the model generates concise answers without reflection. With a larger budget (right), it mirrors full-length reasoning: it starts with problem analysis and using reflective phrases like “wait” and “double-checking.” In both settings, the trace ends appropriately, highlighting our method’s flexibility and controllability.

\section{Conclusion}
\label{sec:conclusion}

We introduce \textit{budget guidance}, a simple yet effective approach for steering LLM reasoning under a thinking budget. Without requiring any LLM fine-tuning, our method enables natural control over the reasoning process and significantly improves token efficiency on challenging benchmarks. Our results demonstrate that LLMs can be effectively guided to reason with \textit{budget guidance}, highlighting budget-conditioned generation as a promising direction for efficient and controllable LLM reasoning.

\bibliography{main}

\begin{thebibliography}{10}

\bibitem{aopsAIME}
{Art of Problem Solving}.
\newblock Aime problems and solutions, n.d.

\bibitem{aopsAMC12}
{Art of Problem Solving}.
\newblock Amc 12 problems and solutions, n.d.

\bibitem{devlin2019bert}
J.~Devlin, M.-W. Chang, K.~Lee, and K.~Toutanova.
\newblock Bert: Pre-training of deep bidirectional transformers for language understanding.
\newblock In {\em Proceedings of the 2019 conference of the North American chapter of the association for computational linguistics: human language technologies, volume 1 (long and short papers)}, pages 4171--4186, 2019.

\bibitem{dhariwal2021diffusion}
P.~Dhariwal and A.~Nichol.
\newblock Diffusion models beat gans on image synthesis.
\newblock {\em Advances in neural information processing systems}, 34:8780--8794, 2021.

\bibitem{dugas2000incorporating}
C.~Dugas, Y.~Bengio, F.~B{\'e}lisle, C.~Nadeau, and R.~Garcia.
\newblock Incorporating second-order functional knowledge for better option pricing.
\newblock {\em Advances in neural information processing systems}, 13, 2000.

\bibitem{openr1}
H.~Face.
\newblock Open r1: A fully open reproduction of deepseek-r1, January 2025.

\bibitem{guo2025deepseek}
D.~Guo, D.~Yang, H.~Zhang, J.~Song, R.~Zhang, R.~Xu, Q.~Zhu, S.~Ma, P.~Wang, X.~Bi, et~al.
\newblock Deepseek-r1: Incentivizing reasoning capability in llms via reinforcement learning.
\newblock {\em arXiv preprint arXiv:2501.12948}, 2025.

\bibitem{han2022folio}
S.~Han, H.~Schoelkopf, Y.~Zhao, Z.~Qi, M.~Riddell, W.~Zhou, J.~Coady, D.~Peng, Y.~Qiao, L.~Benson, et~al.
\newblock Folio: Natural language reasoning with first-order logic.
\newblock {\em arXiv preprint arXiv:2209.00840}, 2022.

\bibitem{han2024token}
T.~Han, Z.~Wang, C.~Fang, S.~Zhao, S.~Ma, and Z.~Chen.
\newblock Token-budget-aware llm reasoning.
\newblock {\em arXiv preprint arXiv:2412.18547}, 2024.

\bibitem{hao2024training}
S.~Hao, S.~Sukhbaatar, D.~Su, X.~Li, Z.~Hu, J.~Weston, and Y.~Tian.
\newblock Training large language models to reason in a continuous latent space.
\newblock {\em arXiv preprint arXiv:2412.06769}, 2024.

\bibitem{he2024olympiadbench}
C.~He, R.~Luo, Y.~Bai, S.~Hu, Z.~L. Thai, J.~Shen, J.~Hu, X.~Han, Y.~Huang, Y.~Zhang, et~al.
\newblock Olympiadbench: A challenging benchmark for promoting agi with olympiad-level bilingual multimodal scientific problems.
\newblock {\em arXiv preprint arXiv:2402.14008}, 2024.

\bibitem{hendrycks2021measuring}
D.~Hendrycks, C.~Burns, S.~Kadavath, A.~Arora, S.~Basart, E.~Tang, D.~Song, and J.~Steinhardt.
\newblock Measuring mathematical problem solving with the math dataset.
\newblock {\em arXiv preprint arXiv:2103.03874}, 2021.

\bibitem{ho2022classifier}
J.~Ho and T.~Salimans.
\newblock Classifier-free diffusion guidance.
\newblock {\em arXiv preprint arXiv:2207.12598}, 2022.

\bibitem{hou2025thinkprune}
B.~Hou, Y.~Zhang, J.~Ji, Y.~Liu, K.~Qian, J.~Andreas, and S.~Chang.
\newblock Thinkprune: Pruning long chain-of-thought of llms via reinforcement learning.
\newblock {\em arXiv preprint arXiv:2504.01296}, 2025.

\bibitem{jaech2024openai}
A.~Jaech, A.~Kalai, A.~Lerer, A.~Richardson, A.~El-Kishky, A.~Low, A.~Helyar, A.~Madry, A.~Beutel, A.~Carney, et~al.
\newblock Openai o1 system card.
\newblock {\em arXiv preprint arXiv:2412.16720}, 2024.

\bibitem{jain2024livecodebench}
N.~Jain, K.~Han, A.~Gu, W.-D. Li, F.~Yan, T.~Zhang, S.~Wang, A.~Solar-Lezama, K.~Sen, and I.~Stoica.
\newblock Livecodebench: Holistic and contamination free evaluation of large language models for code.
\newblock {\em arXiv preprint arXiv:2403.07974}, 2024.

\bibitem{li2024numinamath}
J.~Li, E.~Beeching, L.~Tunstall, B.~Lipkin, R.~Soletskyi, S.~Huang, K.~Rasul, L.~Yu, A.~Q. Jiang, Z.~Shen, et~al.
\newblock Numinamath: The largest public dataset in ai4maths with 860k pairs of competition math problems and solutions.
\newblock {\em Hugging Face repository}, 13:9, 2024.

\bibitem{lightman2023let}
H.~Lightman, V.~Kosaraju, Y.~Burda, H.~Edwards, B.~Baker, T.~Lee, J.~Leike, J.~Schulman, I.~Sutskever, and K.~Cobbe.
\newblock Let's verify step by step.
\newblock In {\em The Twelfth International Conference on Learning Representations}, 2023.

\bibitem{ma2025reasoning}
W.~Ma, J.~He, C.~Snell, T.~Griggs, S.~Min, and M.~Zaharia.
\newblock Reasoning models can be effective without thinking.
\newblock {\em arXiv preprint arXiv:2504.09858}, 2025.

\bibitem{muennighoff2025s1}
N.~Muennighoff, Z.~Yang, W.~Shi, X.~L. Li, L.~Fei-Fei, H.~Hajishirzi, L.~Zettlemoyer, P.~Liang, E.~Cand{\`e}s, and T.~Hashimoto.
\newblock s1: Simple test-time scaling.
\newblock {\em arXiv preprint arXiv:2501.19393}, 2025.

\bibitem{qi2023fine}
X.~Qi, Y.~Zeng, T.~Xie, P.-Y. Chen, R.~Jia, P.~Mittal, and P.~Henderson.
\newblock Fine-tuning aligned language models compromises safety, even when users do not intend to!
\newblock {\em arXiv preprint arXiv:2310.03693}, 2023.

\bibitem{rein2024gpqa}
D.~Rein, B.~L. Hou, A.~C. Stickland, J.~Petty, R.~Y. Pang, J.~Dirani, J.~Michael, and S.~R. Bowman.
\newblock Gpqa: A graduate-level google-proof q\&a benchmark.
\newblock In {\em First Conference on Language Modeling}, 2024.

\bibitem{willard2023efficient}
B.~T. Willard and R.~Louf.
\newblock Efficient guided generation for large language models.
\newblock {\em arXiv preprint arXiv:2307.09702}, 2023.

\bibitem{wu2025tablebench}
X.~Wu, J.~Yang, L.~Chai, G.~Zhang, J.~Liu, X.~Du, D.~Liang, D.~Shu, X.~Cheng, T.~Sun, et~al.
\newblock Tablebench: A comprehensive and complex benchmark for table question answering.
\newblock In {\em Proceedings of the AAAI Conference on Artificial Intelligence}, volume~39, pages 25497--25506, 2025.

\bibitem{qwen2}
A.~Yang, B.~Yang, B.~Hui, B.~Zheng, B.~Yu, C.~Zhou, C.~Li, C.~Li, D.~Liu, F.~Huang, G.~Dong, H.~Wei, H.~Lin, J.~Tang, J.~Wang, J.~Yang, J.~Tu, J.~Zhang, J.~Ma, J.~Xu, J.~Zhou, J.~Bai, J.~He, J.~Lin, K.~Dang, K.~Lu, K.~Chen, K.~Yang, M.~Li, M.~Xue, N.~Ni, P.~Zhang, P.~Wang, R.~Peng, R.~Men, R.~Gao, R.~Lin, S.~Wang, S.~Bai, S.~Tan, T.~Zhu, T.~Li, T.~Liu, W.~Ge, X.~Deng, X.~Zhou, X.~Ren, X.~Zhang, X.~Wei, X.~Ren, Y.~Fan, Y.~Yao, Y.~Zhang, Y.~Wan, Y.~Chu, Y.~Liu, Z.~Cui, Z.~Zhang, and Z.~Fan.
\newblock Qwen2 technical report.
\newblock {\em arXiv preprint arXiv:2407.10671}, 2024.

\bibitem{qwen2.5}
A.~Yang, B.~Yang, B.~Zhang, B.~Hui, B.~Zheng, B.~Yu, C.~Li, D.~Liu, F.~Huang, H.~Wei, H.~Lin, J.~Yang, J.~Tu, J.~Zhang, J.~Yang, J.~Yang, J.~Zhou, J.~Lin, K.~Dang, K.~Lu, K.~Bao, K.~Yang, L.~Yu, M.~Li, M.~Xue, P.~Zhang, Q.~Zhu, R.~Men, R.~Lin, T.~Li, T.~Xia, X.~Ren, X.~Ren, Y.~Fan, Y.~Su, Y.~Zhang, Y.~Wan, Y.~Liu, Z.~Cui, Z.~Zhang, and Z.~Qiu.
\newblock Qwen2.5 technical report.
\newblock {\em arXiv preprint arXiv:2412.15115}, 2024.

\bibitem{yu2025z1}
Z.~Yu, Y.~Wu, Y.~Zhao, A.~Cohan, and X.-P. Zhang.
\newblock Z1: Efficient test-time scaling with code.
\newblock {\em arXiv preprint arXiv:2504.00810}, 2025.

\end{thebibliography}
\bibliographystyle{abbrv}

\appendix

\section*{Appendix}

\section{Quantitative Reasoning Behavior Analysis}

\begin{figure}[h]
    \centering
    \includegraphics[width=\textwidth]{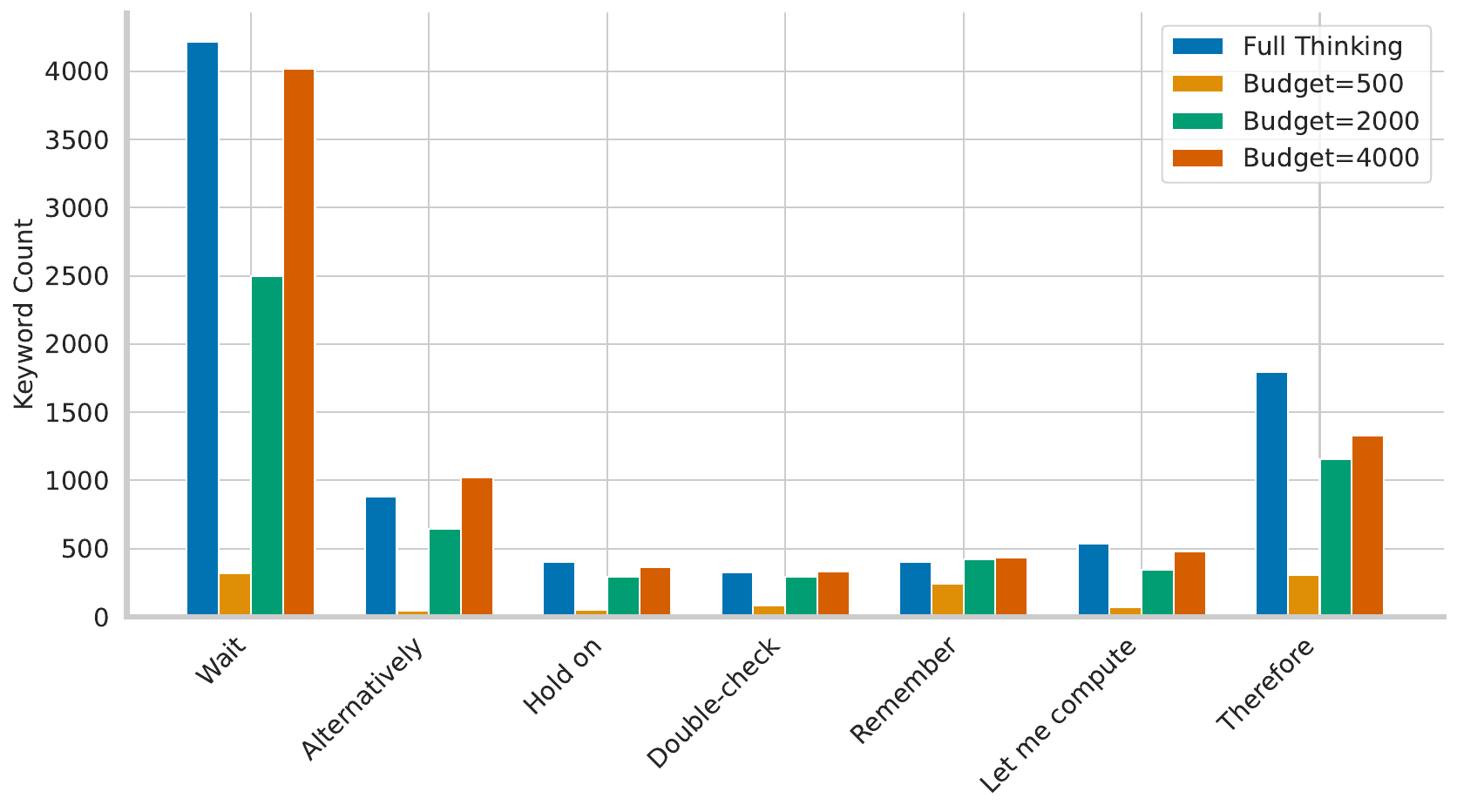}
    \caption{Reasoning keywords frequency comparison under different budget settings.}
    \label{fig:reasoning_word_freq}
\end{figure}

To quantitatively analyze how the predictor influences the reasoning behavior of LLMs under different budget settings, we follow the methodology proposed by~\citep{hou2025thinkprune}. Specifically, we count the frequency of reasoning-related keywords such as \textit{``wait''} and \textit{``alternatively''}, which are indicative of deeper reasoning processes. We compare the keyword frequencies for thinking budget of 500, 2000, and 4000 tokens using the DS-7B model on the MATH-500 benchmark. These results are contrasted with a full-thinking baseline (\textit{i.e.}, without applying our method). The comparison is illustrated in Figure~\ref{fig:reasoning_word_freq}.

As shown in the figure, a smaller budget substantially reduces the frequency of reasoning-related keywords, indicating a more concise reasoning process. As the budget increases, the model is encouraged to engage in deeper reasoning. Notably, when the budget is set sufficiently high, the behavior closely matches that of the full-thinking baseline, suggesting minimal loss in reasoning capability. These findings demonstrate that our method can effectively steer the reasoning behavior of LLMs, while still preserving their reasoning ability under higher budget constraints.

\section{Additional Reasoning Samples}

We provide additional reasoning samples to facilitate a clearer understanding of our method. We selected illustrative examples from MATH-500. Compared to \textit{budget forcing}~\citep{muennighoff2025s1}, our method demonstrates \textbf{more precise and rational thinking length control}. Under identical budget constraints, our method successfully solves example problems where budget forcing fails. 

\subsection{Comparative Analysis: Budget Forcing vs. Budget Guidance}

\begin{figure}[h]
    \centering
    \includegraphics[width=\textwidth]{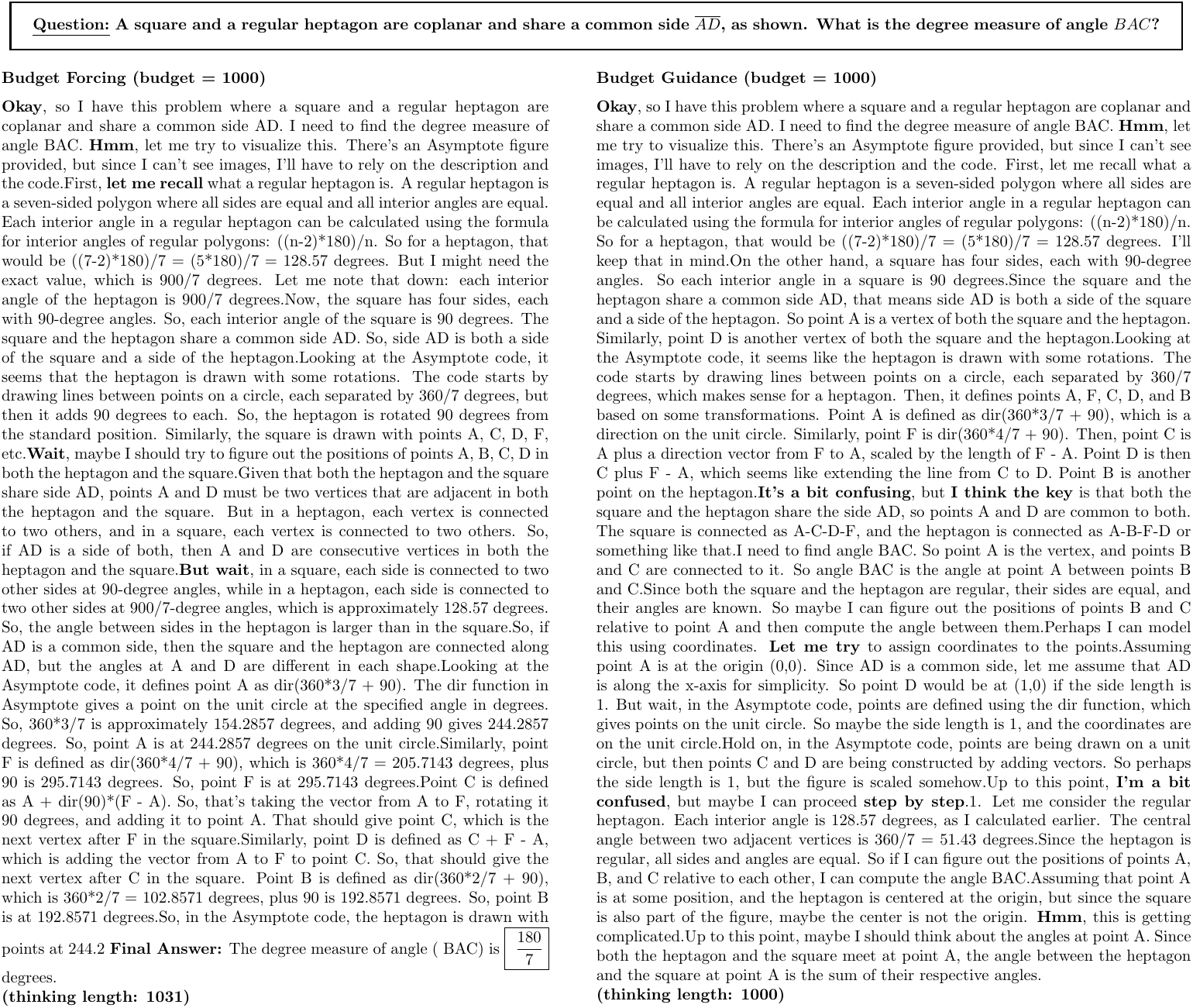}
    \caption{Budget forcing and budget guidance generate a similar number of thinking tokens under a specified thinking budget, but only budget guidance leads to the correct answer.}
    \label{fig:case1}
\end{figure}

\begin{figure}[t]
    \centering
    \includegraphics[width=\textwidth]{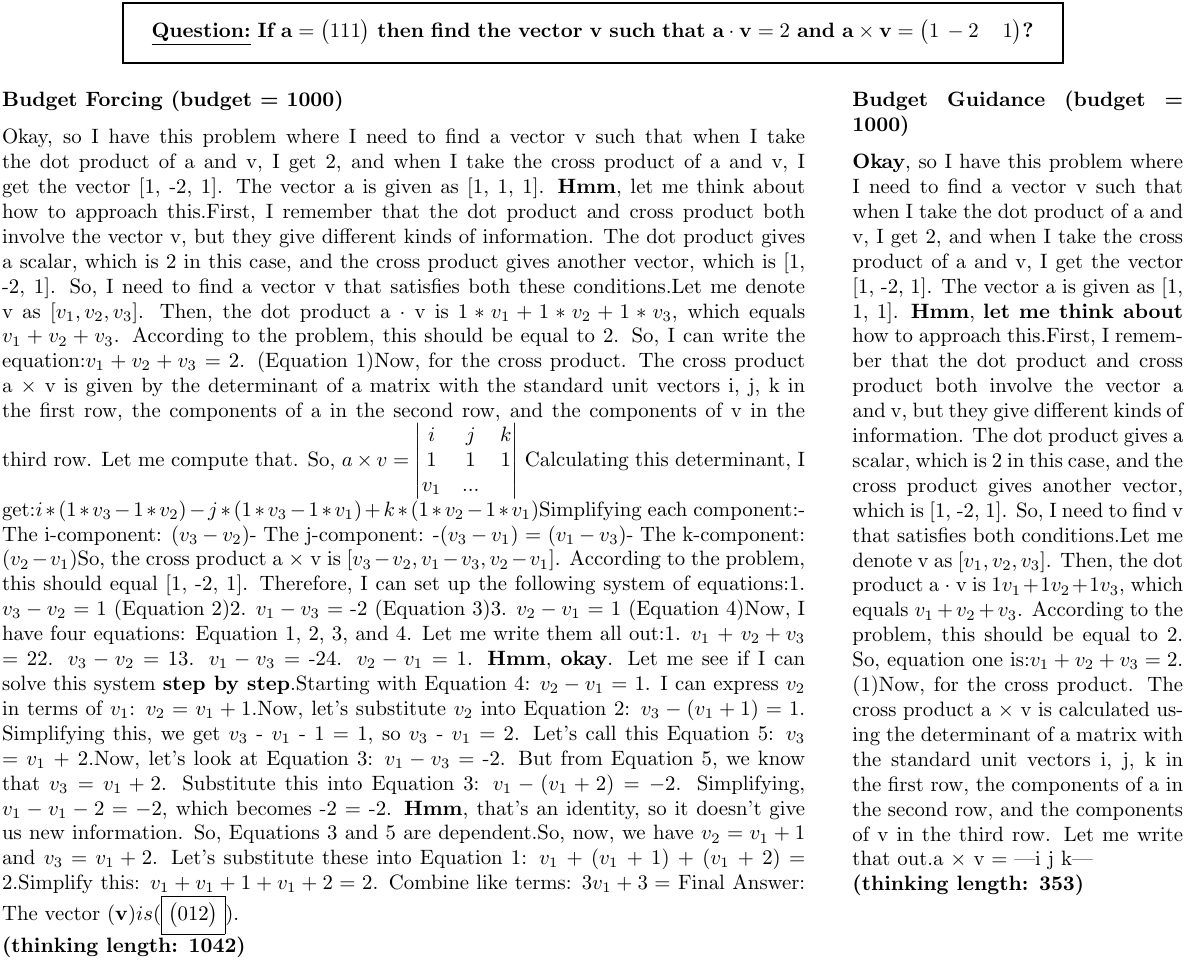}
    \caption{Another example demonstrating that budget guidance can solve a question correctly using fewer thinking tokens.}
    \label{fig:case2}
\end{figure}

We first conduct a comparison between our baseline, \textit{budget forcing}, and our method, \textit{budget guidance}, under the same thinking budget. Figure \ref{fig:case1} and \ref{fig:case2} present cases with a 1000-token thinking budget: The first case in Figure~\ref{fig:case1} shows our method precisely controlling thinking length to just under this limit, while the second case in Figure~\ref{fig:case2} illustrates a drastically shortened reasoning process that still achieves accurate solution. These examples collectively demonstrate the superior performance of our method in effectively controlling the reasoning length within the specified thinking budget, while consistently arriving at the correct answer regardless of problem difficulty.

\subsection{Reasoning Behavior Under Varying Budgets}

\begin{figure}[t]
    \centering
    \includegraphics[width=\textwidth]{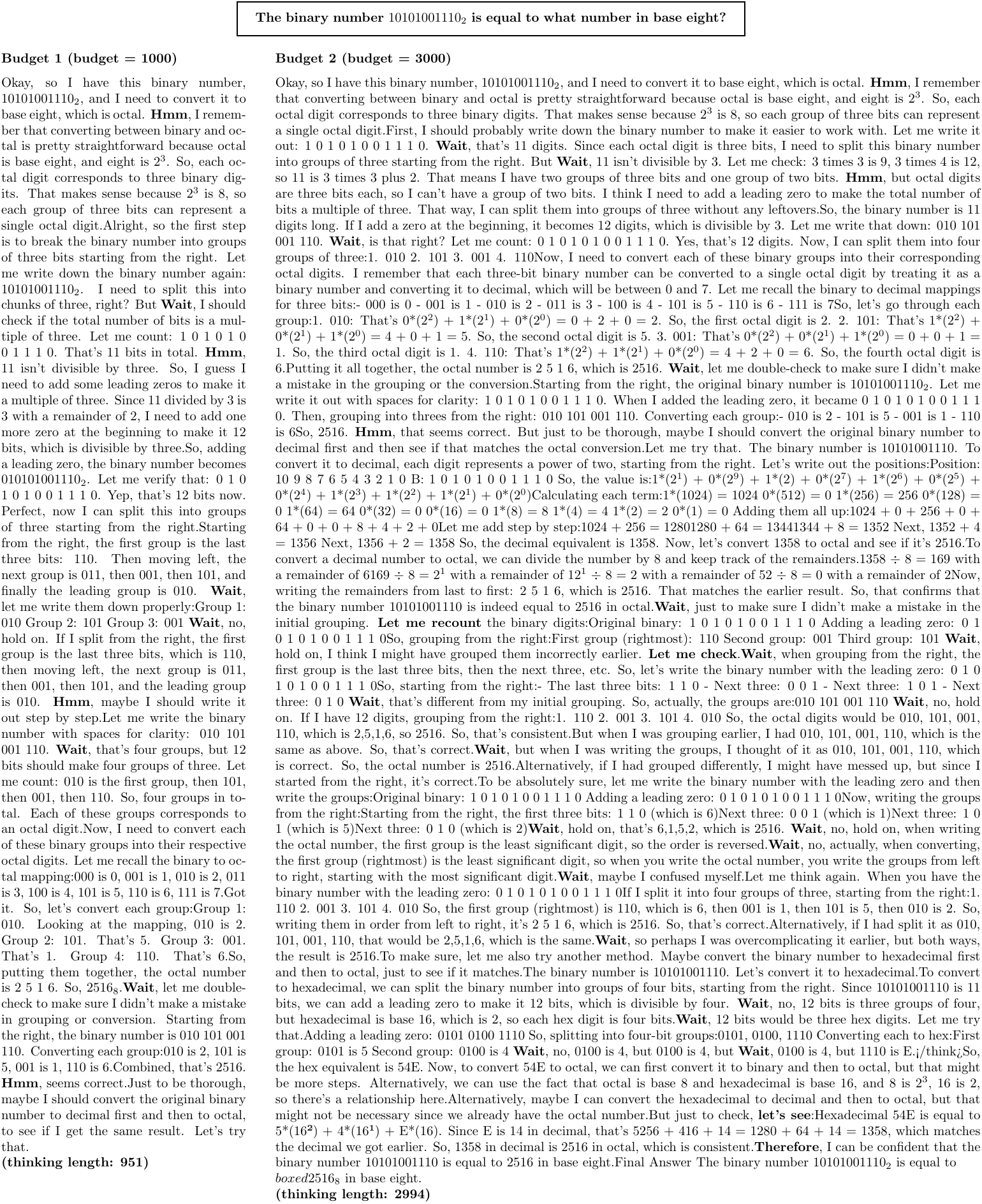}
    \caption{Another example illustrating how budget guidance effectively steers LLM reasoning.}
    \label{fig:case3}
\end{figure}

We further demonstrate our method's ability to \textbf{control thinking length under varying budget constraints}. The results, illustrated in Figure \ref{fig:case3}, show that our method performs excellently across diverse budget settings, flexibly steering the reasoning while ensuring accurate reasoning process.

\section{Dataset Description}

We provide detailed information about the evaluation datasets used in our paper.

\noindent\textbf{MATH-500}~\citep{hendrycks2021measuring} is a 500-problem subset of the MATH dataset, selected by~\citep{lightman2023let}. Each problem is labeled with a difficulty level from 1 to 5.

\noindent\textbf{AIME-2024}~\citep{aopsAIME} contains 30 problems from the 2024 American Invitational Mathematics Examination, covering topics such as algebra, combinatorics, geometry, number theory, and probability. Following \textit{budget forcing}~\citep{muennighoff2025s1}, we retain only the essential ASY figure code required to solve each problem, omitting non-essential diagrams.

\noindent\textbf{AMC}~\citep{aopsAMC12} includes all 83 problems from AMC12 2022 and AMC12 2023.

\noindent\textbf{OlympiadBench}~\citep{he2024olympiadbench} is a challenging benchmark aimed at advancing AGI through Olympiad-level, bilingual, multimodal scientific problems. We use its math subset, which contains a total of 675 problems.

\noindent\textbf{GPQA Diamond}~\citep{rein2024gpqa} consists of 198 high-quality, extremely difficult questions spanning a broad range of scientific domains, including biology, physics, and chemistry.

\noindent\textbf{FOLIO}~\citep{han2022folio} is a human-annotated dataset designed to evaluate complex logical reasoning in natural language. It features 1,430 unique conclusions paired with 487 sets of premises, all validated using first-order logic (FOL) annotations. We use the test set, which contains 203 unique problems.

\noindent\textbf{TableBench}~\citep{wu2025tablebench} is a benchmark for evaluating LLMs on real-world tabular data challenges. We evaluate all models on the numerical reasoning subset, which comprises 493 problems.

\noindent\textbf{LiveCodeBench}~\citep{jain2024livecodebench} offers a holistic and contamination-free evaluation of LLM coding capabilities. Following~\citep{guo2025deepseek}, we select problems from the August 2024 to January 2025 period, totaling 323 problems.

\section{Training Data Augmentation}

We adopt a simple data augmentation strategy to double the size of the training set. Each training sample originally follows the format:
\begin{equation}
\texttt{\textless{}think\textgreater{}THINK\_MESSAGE\textless{}/think\textgreater{}ANSWER\_MESSAGE}    
\end{equation}

Since our predictor only operates on the \texttt{THINK\_MESSAGE}, the \texttt{ANSWER\_MESSAGE} is not used during training. To utilize this otherwise unused information, we construct an additional training sample in the following format:
\begin{equation}
\texttt{\textless{}think\textgreater{}ANSWER\_MESSAGE\textless{}/think\textgreater{}ANSWER\_MESSAGE}
\end{equation}

This transformation allows us to incorporate the \texttt{ANSWER\_MESSAGE} into the predictor's training process. By generating one new sample for each original sample, we effectively double the size of the training set and ensure full utilization of the available data.

\section{Prompt Description}

In Section ~\ref{sec:difficulty-proxy}, we analyze the predictor’s estimated thinking length across different prompt types to demonstrate its prompt awareness. Below, we list the specific prompts used in our experiment.

The prompt for long reasoning is: \texttt{Think step by step and provide thorough reasoning before reaching a conclusion.}

The prompt for short reasoning is: \texttt{Think quickly and provide a concise reasoning with minimal steps.}

We add these prompts as the system prompt.

\end{document}